%% file: main.tex
\pdfoutput=1

\documentclass[11pt]{article}

\usepackage[]{acl}

\usepackage{times}
\usepackage{latexsym}

\usepackage[T1]{fontenc}

\usepackage[utf8]{inputenc}

\usepackage{microtype}

\usepackage{inconsolata}

\usepackage{graphicx}
\usepackage{amsmath}
\usepackage{amssymb}
\usepackage[T1]{fontenc}
\usepackage{graphicx}
\usepackage{latexsym}
\usepackage{xspace}
\usepackage{caption}
\usepackage{subcaption} 
\usepackage{cprotect}
\usepackage{hhline}
\usepackage{array}
\usepackage{float}
\usepackage{xcolor}
\usepackage[group-separator={,},output-decimal-marker = {.}, group-minimum-digits={3}]{siunitx}  
\usepackage{colortbl}
\usepackage{booktabs} 
\usepackage{multirow}
\usepackage[shortlabels]{enumitem} 
\usepackage{tablefootnote}
\usepackage{arydshln}
\usepackage[capitalize]{cleveref}
\crefname{section}{Sec.}{Secs.}
\Crefname{section}{Section}{Sections}
\Crefname{table}{Table}{Tables}
\crefname{table}{Tab.}{Tabs.}
\newcommand{\ie}{i.e.\xspace}
\newcommand{\eg}{e.g.\xspace}

\newcommand{\vl}{{V\&L}\xspace}
\newcommand{\mypar}[1]{\vspace{0.5em}\noindent\textbf{#1}.}
\newcommand{\ourapp}{{\sc {II-MMR}}\xspace}
\newcommand{\ourappbf}{{\textbf{\textsc {II-MMR}}}\xspace}

\newcommand{\apcot}{{\sc {II-MMR\textsubscript{ApCoT}}}\xspace}
\newcommand{\apcotbf}{{\textbf{\textsc {II-MMR\textsubscript{ApCoT}}}}\xspace}

\newcommand{\apcotgt}{{\sc {II-MMR\textsubscript{ApCoT-GT}}}\xspace}

\newcommand{\ktprompt}{{\sc {II-MMR\textsubscript{KtPrompt}}}\xspace}

\newif\ifdraft
\drafttrue
\draftfalse
\ifdraft
  \newcommand{\jihyung}[1]{{\color{red}Jihyung: #1}\xspace}
  \newcommand{\dk}[1]{{\color{blue}DK: #1}\xspace}
  \newcommand{\jk}[1]{{\color{brown}JK: #1}\xspace}
  \usepackage{soul}
\else
  \newcommand{\jihyung}[1]{}
  \newcommand{\dk}[1]{}
  \newcommand{\jk}[1]{}
\fi

\title{\ourappbf: Identifying and Improving \\ Multi-modal Multi-hop Reasoning in Visual Question Answering}

\author{Jihyung Kil$^1$\thanks{\quad Work was partially done during the Amazon internship.}$\footnotemark[1]$ \quad {Farideh Tavazoee$^2$} \quad
{Dongyeop Kang$^3$} \quad {Joo-Kyung Kim$^2$} \vspace{0.5em} \\
{$^1$The Ohio State University} \quad {$^2$Amazon AGI} \quad {$^3$University of Minnesota} \\
{\tt\small\ kil.5@osu.edu} {\tt\small\{fayt,jookyk\}@amazon.com} {\tt\small\ dongyeop@umn.edu}}

\begin{document}
\maketitle
\input{abs}

\input{intro}
\input{approach}
\input{exp}

\input{related}
\input{conclusion}
\input{limitations}
\newpage

\bibliography{anthology,custom}
\input{supp_content}


\end{document}

%% file: abs.tex
\begin{abstract}
Visual Question Answering (VQA) often involves diverse reasoning scenarios across Vision and Language (\vl). Most prior VQA studies, however, have merely focused on assessing the model's overall accuracy without evaluating it on different reasoning cases.
Furthermore, some recent works observe that conventional Chain-of-Thought (CoT) prompting fails to generate effective reasoning for VQA, especially for complex scenarios requiring multi-hop reasoning.
In this paper, we propose \ourappbf, a novel idea to \textbf{i}dentify and \textbf{i}mprove \textbf{m}ulti-modal \textbf{m}ulti-hop \textbf{r}easoning in VQA.
In specific, \ourapp takes a VQA question with an image and finds a reasoning path to reach its answer using two novel language promptings: (i) answer prediction-guided CoT prompt, or (ii) knowledge triplet-guided prompt. 
\ourapp then analyzes this path to identify different reasoning cases 
in current VQA benchmarks by estimating \textbf{how many hops} and \textbf{what types} (\ie, visual or beyond-visual) of reasoning are required to answer the question.
On popular benchmarks including GQA and A-OKVQA, \ourapp observes that most of their VQA questions are easy to answer, simply demanding ``single-hop'' reasoning,
whereas only a few questions require ``multi-hop'' reasoning.
Moreover, while recent \vl models struggle with such complex multi-hop reasoning questions even using the traditional CoT method,~\ourapp shows its effectiveness across all reasoning cases in both zero-shot and fine-tuning settings.\footnote{\url{https://github.com/heendung/II-MMR}}
\end{abstract}

%% file: intro.tex
\section{Introduction}
\label{sec:intro}

\begin{figure}
    \centering
    \centerline{\includegraphics[width=1.0\linewidth]{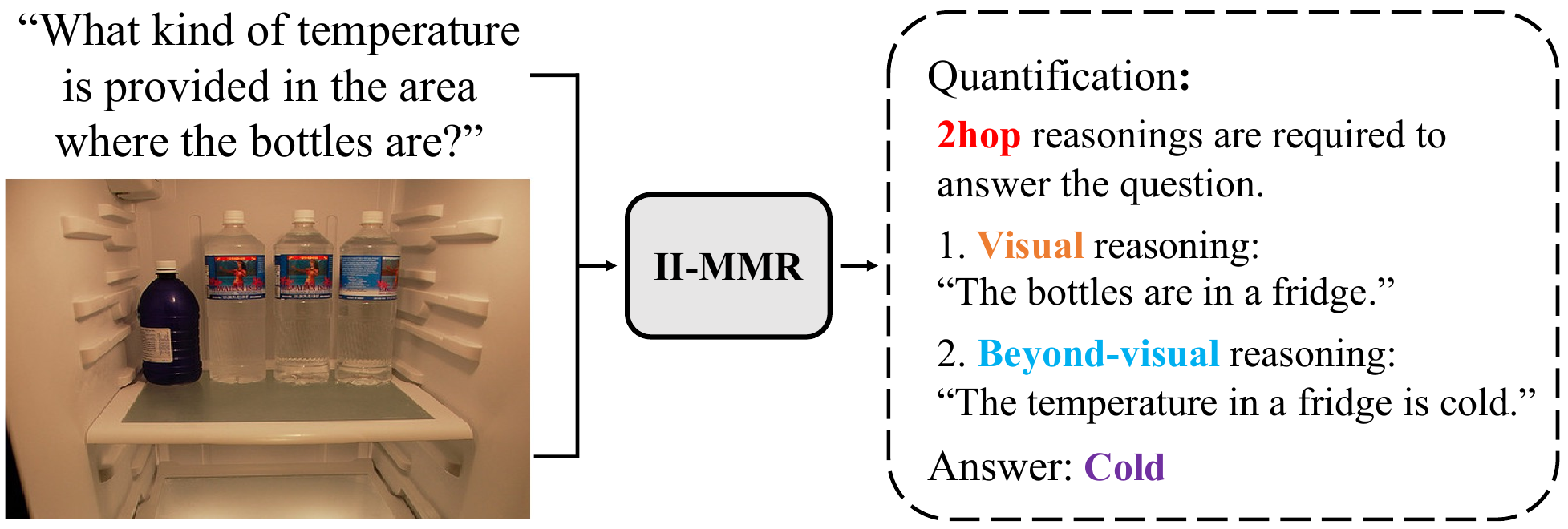}}
    \caption{\small \textbf{Overview of \ourappbf.}  Our \ourapp automatically identifies different reasoning cases in VQA benchmarks by measuring {\color{red}how many} and what types ({\color{orange} {visual}} or {\color{cyan} beyond-visual}) of reasoning are required to solve a VQA question. The identified reasoning process in \ourapp also helps make a correct prediction ({\color{violet}Cold}), while the simple Chain-of-Thought (CoT) method~\cite{kojima2022large} fails to answer.}
    \label{fig:overview}
\end{figure}

Reasoning is a key aspect of highly intelligent systems. 
Visual question answering (VQA)~\cite{goyal2017making,aokvqa,gqa} enables us to measure such reasoning ability as it contains different reasoning scenarios in the benchmark. For instance, a VQA question \emph{``What color is the banana?''} requires one-hop (one-step) reasoning to be answered, which is to identify the color of the banana in the image. In contrast, the other question, \emph{``Which American president is associated with the stuffed animal seen here?''} asks two-hop reasoning:~(i) visually detecting this animal as ``Teddy bear'', and (ii) knowing that the ``Teddy bear'' is related to  President ``Roosevelt'' (\ie, commonsense reasoning).

Despite different reasoning approaches being required for different questions, most prior VQA studies~\cite{tan2019lxmert,chen2022pali,wang2022git} have solely focused on the model's \emph{overall} accuracy, neglecting to evaluate its reasoning capabilities. 
While a few works~\cite{li2018vqa,wu2019self} attempt to interpret its reasoning process, they often rely on human explanations, which are challenging to collect sufficiently. Moreover, most VQA benchmarks~\cite{goyal2017making,aokvqa} do not provide detailed information on reasoning, including how many and what types of reasoning are required to answer the question. These limitations hinder the extensive and in-depth understanding of the model's reasoning abilities.
Recently, Chain-of-Thought (CoT) prompting~\cite{wei2022chain,kojima2022large}, which elicits complex multi-hop reasoning via step-by-step instructions, has demonstrated remarkable performance across various NLP domains, including arithmetic and logical reasoning~\cite{cobbe2021training,srivastava2023beyond}. 
However, some recent studies~\cite{awal2023investigating} point out that this CoT reasoning may be ineffective for VQA tasks due to (i) the model's limited capabilities to ground visual objects in rationale generation and (ii) its proneness to hallucinate non-existent objects in the image.

\textbf{Proposal.}
In this paper, we propose \ourappbf, a novel method which automatically \textbf{i}dentifies and \textbf{i}mproves \textbf{m}ulti-modal \textbf{m}ulti-hop \textbf{r}easoning for VQA tasks. 
Given a VQA question with its relevant image, our \ourapp first identifies a reasoning path to reach its answer using two novel prompting strategies:~(i) answer prediction-guided CoT (\apcot), or~(ii) knowledge triplet-guided prompt (\ktprompt). 
Concretely, \apcot first prompts a \vl model to directly predict an answer for the question and then generates an answer-related path by incorporating this prediction into the CoT prompt, guiding reasoning towards the answer. Besides, \ktprompt asks an LLM to extract knowledge triplets from question and answer (QA) and treats the sequence of these triplets as the answer reasoning path. 
In short, \ourapp utilizes additional cues, either through answer prediction or QA-related knowledge triplets, to find the correct reasoning path.

\textbf{Effectiveness of \ourapp.}
\ourapp 
analyzes this reasoning path to identify different reasoning cases in current VQA benchmarks by measuring the number of reasoning steps and the types of reasoning, such as visual or beyond-visual (\eg, commonsense, knowledge base \cite{aokvqa}), required to answer the question (\autoref{fig:overview}). 
During our prompting process, the intermediate reasoning steps and the alignment of question keywords with visual objects determine the number and types of reasoning.

\ourapp finds two shortcomings of GQA~\cite{gqa} and A-OKVQA~\cite{aokvqa} benchmarks: 
(i) a scarcity of multi-hop reasoning questions and (ii) an overestimation of VQA performance due to the high model accuracy on simple one-hop reasoning questions. 
Concretely, while the current well-known \vl model (\eg, BLIP-2~\citet{blip2}) excels in such one-hop reasoning questions, it struggles in complex multi-hop scenarios, even using the standard CoT reasoning~\cite{kojima2022large}. 
On the other hand, \ourapp with the proposed language promptings shows notable performance across all reasoning scenarios, including multi-hop cases in both zero-shot and fine-tuning settings.

In short, our \ourapp suggests that identifying (or breaking down) reasoning helps a better understanding of the internal reasoning process and improves the reasoning performance in multi-hop scenarios. Moreover, we believe our \ourapp could be used to create a more complex and practical multi-hop VQA dataset for future work.

\begin{figure}
    \centering
    \centerline{\includegraphics[width=1.0\linewidth]{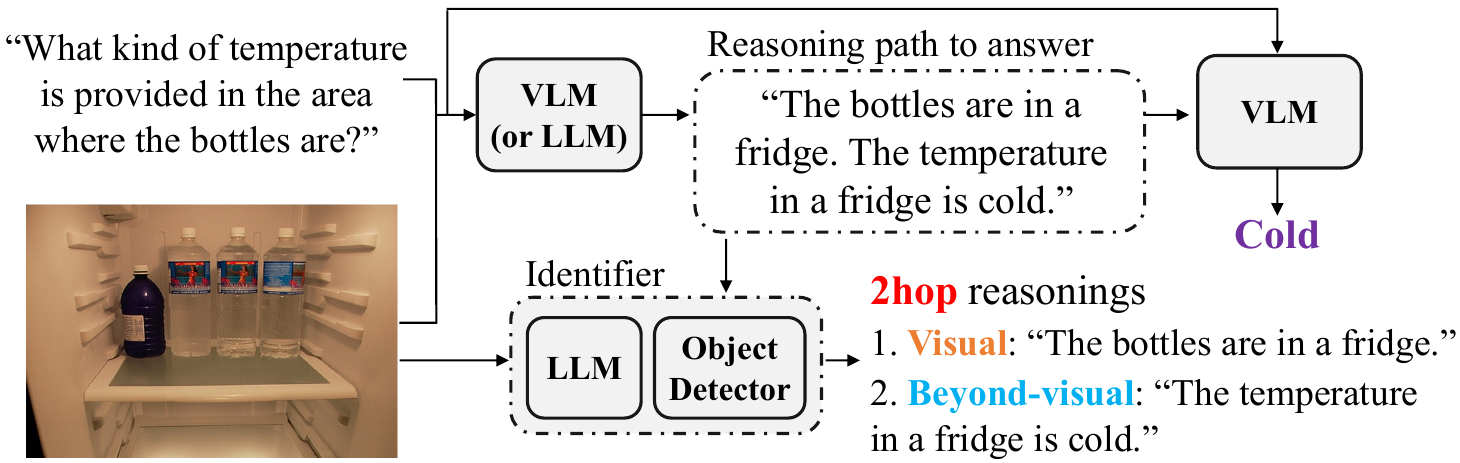}}
    \caption{\small \textbf{Pipeline of \ourappbf.}  Given a VQA question with its image, \ourapp first generates a reasoning path to the answer either using the \vl model (VLM) or the LLM. We then utilize this path to identify different reasoning cases in VQA benchmarks by estimating the {\color{red}number} and types ({\color{orange} {visual}} or {\color{cyan} beyond-visual}) of reasoning required for the question. Finally, \ourapp feeds the reasoning path, along with the question and the image, into VLM to predict the {\color{violet} answer}.}
    \label{fig:general_pipeline}
\vspace{-5pt}
\end{figure}

\begin{figure*}
    \centering
    \centerline{\includegraphics[width=0.95\linewidth]{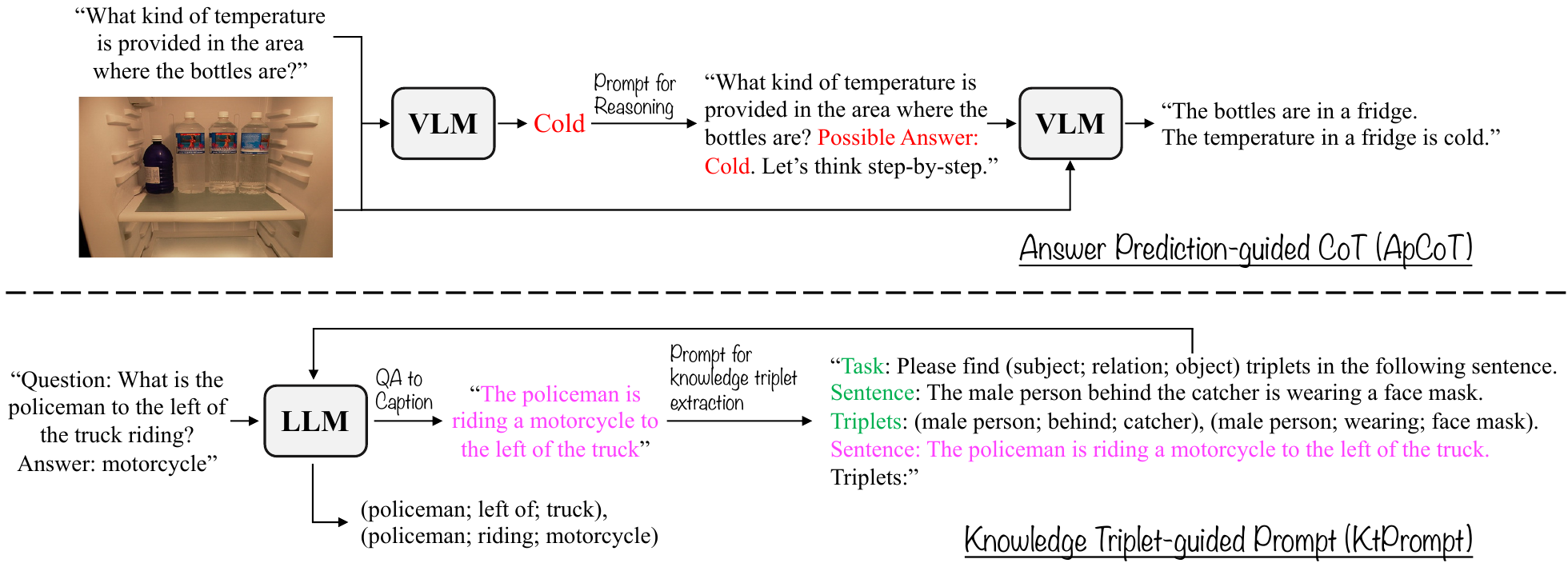}}
    \caption{\small \textbf{The language promptings of our \ourappbf.}  \textbf{Top}: \apcot first asks the VLM to predict an answer for a VQA question. It then integrates its {\color{red} prediction} into the CoT prompt to generate an answer-related rationale, a sequence of reasoning sentences.
    \textbf{Bottom}: \ktprompt initially instructs the LLM to convert the question and answer (QA) to the {\color{magenta} caption}. Then, \ktprompt inputs a prompt (with {\color{teal} task}, {\color{teal} in-context example}, and {\color{magenta} target caption}) to the LLM to extract knowledge triplets from QA. We treat the sequence of sentences (or knowledge triplets) as the reasoning path to reach the answer.}
    \label{fig:lang_prompting}
\end{figure*}

Our main contributions are three-folded:
\begin{itemize} [itemsep=0pt,topsep=0.0pt,leftmargin=5pt,partopsep=0pt]
    \item We introduce \ourapp to identify and improve the multi-hop reasoning for VQA tasks (\autoref{fig:overview}).
    \item \ourapp finds that current VQA benchmarks have some flaws, including the shortage of multi-hop reasoning questions 
    and the inflated results due to simple reasoning cases.   
    \item \ourapp shows its effectiveness in all reasoning cases, including multi-hop reasoning ones in both zero-shot and fine-tuning settings.
\end{itemize}


%% file: approach.tex
\section{Proposed Approach: \ourapp}
\label{sec:approach}

\autoref{fig:general_pipeline} provides a pipeline of \ourapp. In what follows, we first describe our two novel promptings, \apcot and \ktprompt, to find a reasoning path leading to the answer (\S\ref{sec:reasoning_path}). 
Then, we describe how to identify different reasoning cases in current VQA benchmarks (\S\ref{sec:quantification}) using the detected reasoning paths in \S\ref{sec:reasoning_path}. Finally, we discuss how to utilize the reasoning paths to further improve the reasoning performance in zero-shot and fine-tuning stages (\S\ref{sec:eval_zeroshot_finetune}).  

\dk{I wish to include a high-level pipeline figure to describe these three steps for better understanding. I changed the wording a bit clear but I still believe some may not follow well. }
\jk{+1}
\subsection{Finding a reasoning path to the answer}
\label{sec:reasoning_path}
\subsubsection{Preliminary Analysis}
\label{sec:preliminary}

\begin{table}[h]
\centering
\small
\tabcolsep 6.5pt
\renewcommand\arraystretch{1.0}
\begin{tabular}{cc}
\toprule
Model & A-OKVQA \\
\midrule
BLIP-2 & 46.05 \\
BLIP-2+CoT & 36.06 \\
\bottomrule
\end{tabular}
\caption{\small \textbf{Weakness of traditional CoT prompting.} The zero-shot performance of BLIP-2 on A-OKVQA becomes worse when the conventional CoT reasoning is applied.}
\vspace{-10pt}
\label{tab:weak_trad_cot}
\end{table}

One approach to identifying the reasoning path involves utilizing rationales generated by the large language models (LLMs) through CoT prompting. 
However, the conventional CoT prompting~\cite{kojima2022large} may not be as effective for VQA tasks as for NLP tasks. 
Indeed, we have observed that BLIP-2~\cite{blip2}, one of the prominent \vl models, performs significantly worse on A-OKVQA and GQA when employing CoT reasoning, compared to standard prompting (\eg, A-OKVQA: 36.06\% vs.~46.05\% in \autoref{tab:weak_trad_cot}, GQA: 39.08\% vs.~44.63\% in \autoref{tab:effect_ourCot_diff_reasoning}).
Throughout our comprehensive error analyses, we find that \vl models with CoT fail due to incorrectly or irrelevantly generated rationales. Thus, the rationales generated by the conventional CoT may not be suitable for the reasoning path to the answer. 

\subsubsection{Answer prediction-guided CoT (ApCoT)}
\label{sec:apcot}
To find a better reasoning path, we introduce an answer prediction-guided CoT (\apcot), which assists the model in generating more answer-related rationales by providing its initial predictions as input context. Concretely,~\apcot starts by prompting a \vl model to directly predict an answer for the VQA question. It then incorporates this predicted answer into the CoT prompt to generate a rationale (Top in \autoref{fig:lang_prompting}). We empirically see that the context of the initial prediction leads the model to focus on a topic relevant to the answer and generates a more answer-related rationale. We treat this rationale (concretely, a sequence of reasoning sentences) as a reasoning path to the answer. 


Compared to the traditional CoT prompting, our \apcot notably improves the model answer accuracy, indirectly demonstrating the high quality of our rationales (See \S\ref{subsec:cot_zershot} and \S\ref{subsec:cot_finetune} for more details). Additionally, we conducted a human study with 300 randomly selected VQA samples with our generated rationales and showed the high quality of our rationales. Please see \S\ref{subsec:hop_est} for more details.


\subsubsection{Knowledge Triplet-guided Prompt (KtPrompt)}
\label{sec:ktprompt}

In NLP, a knowledge triplet can be viewed as a one-hop (one-step) reasoning. For example, a question, \emph{``Which team does the player named 2015 Diamond Head Classic’s MVP play for?''} requires two reasoning steps, \emph{``Buddy Hield is MVP for Diamond Head Classic''} and \emph{``Buddy Hield plays for Sacramento Kings''}.
This two-step reasoning can be naturally formed into two knowledge triplets, \emph{(Buddy Hield, MVP, Diamond Head Classic)} and \emph{(Buddy Hield, PlayFor, Sacramento Kings)}. 

Built upon this insight, \ktprompt aims to extract knowledge triplets from the question (with the answer) to identify a reasoning path leading to the answer (Bottom in \autoref{fig:lang_prompting}). 
Concretely, \ktprompt first utilizes an LLM (Llama-2-70b~\citet{touvron2023llama}) to convert the combination of question and answer into a natural caption. Next, we construct an in-context prompt to instruct the LLM to extract knowledge triplets from the provided caption. However, the LLM may generate noisy knowledge triplets. For instance, triplets may lack some components (\eg, subject, relation, or/and object) or contain trivial words like stop-words (\eg, ``the''), which are not typically considered as components.
We filter out such noisy samples and obtain a clean set of knowledge triplets. 
For every VQA question, we treat the sequence of knowledge triplets as a reasoning path to its answer.

\subsection{Analyzing a reasoning path}
\label{sec:quantification}
After obtaining the answer reasoning path for each VQA question, we analyze this path to identify different reasoning cases in current VQA benchmarks by automatically measuring the number and types of reasoning required to answer the question. Specifically, we count one reasoning sentence (or one knowledge triplet) in the path as one reasoning step. Besides, the reasoning sentence is categorized into ``visual'' or ``beyond-visual'' (\autoref{fig:quantifier}). We first task an LLM (Llama-2-70b~\cite{touvron2023llama}) with extracting keywords from a reasoning sentence. Concurrently, we input the image into GLIP~\cite{glip}, a phrase-region grounded object detector, to identify objects in the image. We then check how many keywords in the sentence match the visual objects. If \emph{all} keywords match visual objects, the sentence is classified as ``visual'' reasoning, as it only contains knowledge about the image. If \emph{not all} keywords match, we categorize it as ``beyond-visual'' reasoning since it involves additional knowledge (\eg, commonsense) beyond the visual information.

\subsection{Model performance on the reasoning cases}
\label{sec:eval_zeroshot_finetune}
We investigate whether our \apcot and \ktprompt effectively improve the model answer accuracy in all the reasoning cases identified in VQA benchmarks (\S\ref{sec:quantification}).  
Concretely, after obtaining the rationale (or knowledge triplet) through our methods, we prepend them to an answer-triggering prompt (\eg, ``Therefore, short answer:''). This combined prompt (with the image and the question) is then fed into the \vl model to make a prediction for the question in each reasoning case. We explore two settings: (i) zero-shot, where the pre-trained model is not further trained with the downstream VQA benchmarks, and (ii) fine-tuning, which involves utilizing the downstream VQA training data to train the model.

%% file: exp.tex
\section{Experimental Setup}

\begin{figure}
    \centering
    \centerline{\includegraphics[width=0.95\linewidth]{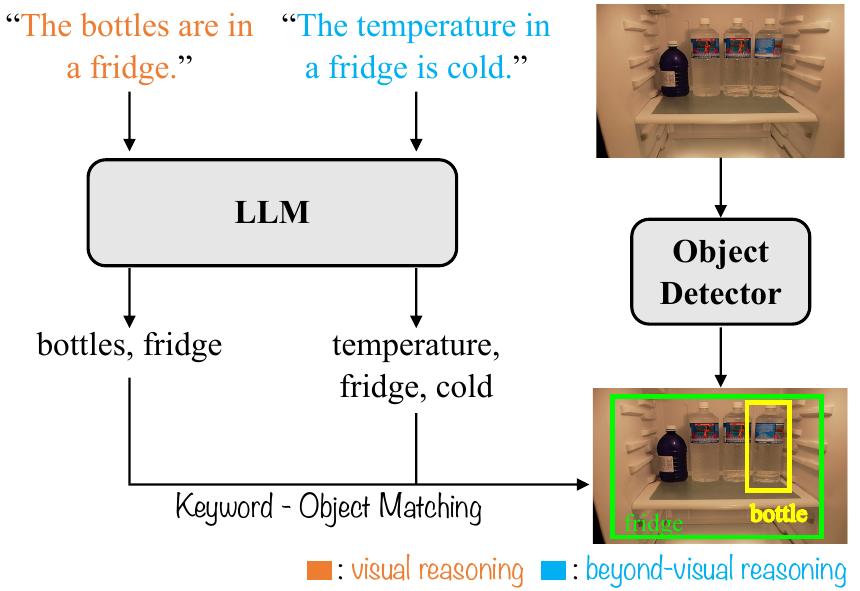}}
    \caption{\small \textbf{Analyzing the reasoning types.} The LLM extracts keywords (\eg, ``bottle'', ``temperature'') from each reasoning sentence. Meanwhile, the object detector identifies objects (\eg, ``fridge'', ``bottle'') in the image. We then check if all keywords match visual objects and decide the reasoning type ({\color{orange} {visual}} or {\color{cyan} beyond-visual}) of each sentence in the rationale.}
    \label{fig:quantifier}
\vspace{-5pt}
\end{figure}

\label{sec:exp}
\mypar{VQA benchmarks} There exists a variety of VQA benchmarks~\cite{antol2015vqa,goyal2017making,marino2019ok,agrawal2018don,aokvqa,gqa}. We explore them in detail and select two VQA benchmarks GQA and A-OKVQA, which most fit with our aim of analyzing different reasoning scenarios. Concretely, GQA is designed to provide compositional reasoning questions over images. A-OKVQA focuses on knowledge-based VQA, requiring knowledge outside images, including commonsense and knowledge base. Due to their design purposes, GQA and A-OKVQA contain multi-hop reasoning questions, useful for analyzing different reasoning cases and understanding the model's reasoning capabilities in various aspects. Please see the appendix for more details about the datasets.

\mypar{Baselines} We mainly conduct our studies with BLIP-2~\cite{blip2}, one of the prominent \vl models equipped with strong zero-shot CoT capabilities. We provide two baselines: BLIP-2 and BLIP-2+CoT, which are models without/with traditional CoT reasoning, respectively.

\mypar{Evaluation metric}
We follow the same evaluation metrics used in GQA~\cite{gqa} and A-OKVQA~\cite{aokvqa}. GQA uses the standard accuracy while the A-OKVQA accuracy is based on the average score over nine subsets of the ground-truth ten answers, where each score is:
$min(\frac{\# answer\ occurrences}{3}, 1)$.

\mypar{Training details}
We follow the official configurations and implementations of BLIP-2\footnote{\url{https://github.com/salesforce/LAVIS/tree/main}}. Specifically, we select the largest BLIP-2 model, which its vision encoder is ViT-g~\cite{fang2023eva} and its LLM is FlanT5\textsubscript{XXL}~\cite{flant5}. 
During fine-tuning, the parameters of BLIP-2 are optimized with the language modeling loss on the downstream VQA training data with a batch size of 16 and a learning rate of 1e-5 for ten epochs.
We utilize eight A100 48GB GPUs for both training and inference. See the appendix for more details.

\mypar{Identifying different reasoning cases}
We note that the number of reasoning steps in GQA is measured based on the ``ground-truth'' reasoning path derived from its scene graph~\cite{krishna2017visual}. GQA questions were generated using scene graph information, such as object relations or attributes. Thus, we leverage this information to obtain the ground-truth path and identify the reasoning cases in GQA, rather than using our generated reasoning path. 
For A-OKVQA, as no ground-truth paths (or scene graph) exist, we rely on the reasoning paths generated by our \apcot. To accurately measure the number of steps, instead of incorporating the model's prediction into the CoT prompt, we include the ground truth answer to generate the answer reasoning path.

\section{Experimenetal Results}
\jk{This section is not very readable. Too many subsections, each of which is quite long. Could you write the most important findings and results here, and let each subsection also specify the highlights at the beginning and then provide the details?}

\mypar{Aim of our experiments}
Our main experimental goals are two-folded: (i) to provide comprehensive statistics on different reasoning cases in current VQA benchmarks (\S\ref{subsec:reasoning_stat}-\S\ref{subsec:hop_est}) and (ii) to conduct a thorough analysis of the performance of our \ourapp in these reasoning cases (\S\ref{subsec:cot_zershot}-\S\ref{subsec:qual_res}).

\subsection{Analysis of reasoning in VQA benchmarks}
\label{subsec:reasoning_stat}

\autoref{tab:gqa_aokvqa_hops} (Top row) shows the distribution of reasoning steps that our \ourapp identified in GQA and A-OKVQA benchmarks. We note that most GQA questions are simple: requiring direct reasoning (0-hop) (48.74\%), involving only the detection of an object in the image (\eg, ``Is this a truck?''), or 1-hop reasoning (47.53\%). Similarly, 1-hop reasoning questions (69.03\%) dominate in A-OKVQA while both benchmarks lack multi-hop reasoning questions (2-hop in GQA: 3.73\%, 2 or more-hops in A-OKVQA: 30.97\%). These indicate that those VQA benchmarks are biased to evaluate the model's reasoning capabilities in \emph{simple} cases.


We further analyze the types of reasoning required in A-OKVQA (cf. \S\ref{sec:quantification}). As shown in \autoref{tab:skwd_vis_text_aokvqa}, the distribution of reasoning type is skewed toward ``beyond-visual'' reasoning, suggesting that many questions require knowledge beyond the image to be answered. This finding aligns with the objective of the A-OKVQA task, which assesses the \vl model's knowledge outside the image, such as commonsense or knowledge bases.


In addition, some zero-shot VQA performances are overestimated by the high accuracies on simple questions (Bottom row in \autoref{tab:gqa_aokvqa_hops}). For instance, 
BLIP-2 severely suffers on multi-hop reasoning (\eg, 7.66\% on 2-hop). 
However, since direct/1-hop reasoning samples dominate in GQA (48.74\%/47.53\%) and the model accuracy on these samples is high (49.62\%/42.41\%), the overall accuracy remains relatively high (44.63\%). This suggests that overall accuracy is inflated by the accuracy of simple questions, and thus, relying solely on the overall accuracy may be insufficient to accurately evaluate the model's reasoning abilities. 
 

\begin{table}[t]
\centering
\small
\tabcolsep 1.5pt
\renewcommand\arraystretch{1.0}
\begin{tabular}{cccccccc}
\toprule
\multirow{2}{*}{Metric} & \multicolumn{4}{c}{GQA} & \multicolumn{3}{c}{A-OKVQA} \\
\cmidrule(r){2-5} \cmidrule(r){6-8}
& 0-hop & 1-hop & 2-hop & All & 1-hop & $\geq$2-hop & All \\
\midrule
Hop & \multirow{2}{*}{48.74} & \multirow{2}{*}{47.53} & \multirow{2}{*}{3.73} & \multirow{2}{*}{100} & \multirow{2}{*}{69.03} & \multirow{2}{*}{30.97} & \multirow{2}{*}{100} \\
Distribution & & & & & & & \\
\midrule
BLIP-2 & \multirow{2}{*}{49.62} & \multirow{2}{*}{42.41} & \multirow{2}{*}{7.66} & \multirow{2}{*}{44.63} & \multirow{2}{*}{46.70} & \multirow{2}{*}{46.54} & \multirow{2}{*}{46.05} \\
Accuracy & & & & & & & \\
\bottomrule
\end{tabular}
\caption{\small \textbf{Hop distribution and model accuracy on GQA and A-OKVQA}. Simple questions (0/1-hop) dominate, while only a few require multi-hop reasoning (\eg, 2-hop). For zero-shot GQA, the overall accuracy is highly biased to the accuracy of ``simple'' questions. In contrast, the model suffers in complex questions requiring multi-hop reasoning.}
\label{tab:gqa_aokvqa_hops}
\end{table}

\begin{table}[t]
\centering
\small
\tabcolsep 15.5pt
\renewcommand\arraystretch{1.0}
\begin{tabular}{ccc}
\toprule
\multirow{2}{*}{Reasoning} & \multicolumn{2}{c}{A-OKVQA} \\
\cmidrule(r){2-3}
Type & 1-hop & 2-hop\\
\midrule
Visual & 37.14 & 36.69\\
Beyond-visual & 62.85 & 63.31\\ 
\bottomrule
\end{tabular}
\caption{\small \textbf{Distribution of reasning type on A-OKVQA.} Most questions require knowledge (\eg, commonsense) beyond visual information, aligned with the purpose of this task.
}
\vspace{-10pt}
\label{tab:skwd_vis_text_aokvqa}
\end{table}

\begin{table}[t]
\centering
\small
\renewcommand\arraystretch{1.0}
\begin{tabular}{ccccc}
\toprule
\multirow{2}{*}{Model} & \multicolumn{4}{c}{GQA} \\
\cmidrule(r){2-5}
& 0-hop & 1-hop & 2-hop & All \\
\midrule
\ktprompt  & 90.55 & 87.26 & 84.04 & 88.74 \\
\bottomrule
\end{tabular}
\caption{\small \textbf{Hop prediction on GQA.}
Our \ktprompt can estimate the number of reasoning steps required to answer questions over all different reasoning cases.}
\label{tab:hop_pred_gqa}
\end{table}

\begin{table}[t]
\centering
\small
\renewcommand\arraystretch{1.0}
\begin{tabular}{ccc}
\toprule
\multirow{2}{*}{Model} & \multicolumn{2}{c}{GQA} \\
\cmidrule(r){2-3}
& Strict Matching & Partial Matching \\
\midrule
\ktprompt & 91.65 & 94.44 \\
\bottomrule
\end{tabular}
\caption{\small \textbf{Accuracy of our reasoning path on GQA.} Our reasoning path generated by \ktprompt highly matches the ground-truth reasoning path, showing its high quality.}
\vspace{-10pt}
\label{tab:reasoning_path_est_gqa}
\end{table}

\begin{table*}[t]
\centering
\small
\tabcolsep 10.5pt
\renewcommand\arraystretch{1.0}
\begin{tabular}{cccccccc}
\toprule
\multirow{2}{*}{Model} &
\multicolumn{4}{c}{GQA} &
\multicolumn{3}{c}{A-OKVQA} \\
\cmidrule(r){2-5} \cmidrule(r){6-8}
& 0-hop & 1-hop & 2-hop & All & 1-hop & $\geq$2-hop & All \\
\midrule
BLIP-2 & 49.62 & 42.41 & 7.66 & 44.63 & 46.70 & 46.54 & 46.05 \\
BLIP-2+CoT & 44.27 & 35.41 & 18.09 & 39.08 & 37.34 & 34.15 & 36.06 \\
BLIP-2+\apcot & 49.69 & 43.88 & 19.36 & 45.79 & \textbf{51.32} & \textbf{48.09} & \textbf{49.31} \\
BLIP-2+\ktprompt & 49.62 & 42.59 & 27.45 & 45.45 & - & - & - \\
\midrule
LLaVA-1.5+CoT & 67.41 & 56.16 & 43.40 & 61.16 & - & - & - \\
LLaVA-1.5+\apcot & \textbf{70.47} & \textbf{57.51} & \textbf{46.60} & \textbf{63.42} & - & - & - \\
\bottomrule
\end{tabular}
\captionsetup{width=0.95\textwidth}
\caption{\small \textbf{Effectiveness of \ourappbf over different reasoning cases in zero-shot VQA.} Our \ourapp outperforms the traditional CoT on every reasoning case. Moreover, \ourapp notably improves over the baseline (BLIP-2) in all reasoning cases on both benchmarks, suggesting the benefit of our answer reasoning path in finding correct answers. Moreover, \ourapp shows its applicability to a recent VLM, LLaVA-1.5~\cite{liu2023llava}. Following \citet{liu2023llava}, we do not evaluate its zero-shot performance on A-OKVQA, as A-OKVQA was used during model training.}
\label{tab:effect_ourCot_diff_reasoning}
\end{table*}

\subsection{Accuracy of predicting hops and reasoning path }
\label{subsec:hop_est}
As mentioned in \S\ref{sec:exp}, for GQA, we are able to obtain the ground-truth reasoning path from its scene graph. Thus, we can evaluate the quality of our reasoning path against the ground-truth one. \autoref{tab:hop_pred_gqa} shows that our \ktprompt is capable of estimating different numbers of reasoning steps required for questions. For instance, \ktprompt correctly predicts the number of hops for 88.74\% of the total samples in GQA. Moreover, we evaluate the correctness of our reasoning path against the ground-truth path using two matching metrics, Strict and Partial. The former ensures all components in each triplet match between our and ground-truth paths, while the latter is a relaxed version, checking if two of the components match. As depicted in \autoref{tab:reasoning_path_est_gqa}, our reasoning path is highly consistent with the ground-truth path, demonstrating the benefit of \ktprompt. 

We note that as the ground-truth GQA path is formatted as a sequence of knowledge triplets, we only compare it with \ktprompt, which shares the same format, rather than \apcot, which utilizes a different format (\ie, the sequence of rationale sentences). For \apcot, we instead conduct a human study to evaluate the quality of its generated rationales. We provide each of the three annotators with 100 A-OKVQA samples along with their rationales and focus on two aspects: (i) the correctness of our rationales and (ii) the correctness of the number of reasoning steps (sentences) within the rationales. The annotators deem our rationales and their number of reasoning steps correct in 82\% and 71\% of samples, respectively (245/214 out of 300), reaffirming the high quality of our rationales.

\subsection{Benefit of \ourappbf in zero-shot stage}
\label{subsec:cot_zershot}
\mypar{Overall Answer Accuracy}
\autoref{tab:effect_ourCot_diff_reasoning} presents the benefit of \ourapp in the context of the model's overall accuracy on GQA and A-OKVQA in the zero-shot setting. 
First, compared to the traditional CoT (BLIP2+CoT), our \apcot shows superior performance (GQA: 39.08\% vs.~45.79\%, A-OKVQA: 36.06\% vs.~49.31\%). Similarly, our \ktprompt outperforms it by 6.3\%. This suggests that providing the prediction (or the knowledge triplet) as input context helps the model correct wrong reasoning paths. 

Second, our \apcot notably outperforms the baseline (BLIP-2) on both benchmarks (\eg, A-OKVQA: 49.31\% vs.~46.05\%), implying that even initially \emph{incorrect} predicted answers are beneficial as input context. Based on the empirical analysis of this case, we find that the incorrect predictions are often closely related to the correct answer. For instance, some questions ask about the object color, and the model indeed predicts the color-related answer but is incorrect (\eg, Prediction: ``purple'', Ground-truth: ``green''). In this case, our \apcot, providing the prediction as a \emph{possible} answer (cf. Top in~\autoref{fig:lang_prompting}), may guide the model to ``rethink'' the correct answer (\eg, true color), eventually fixing the wrong answer. See the appendix for its qualitative examples.

\mypar{Answer Accuracy over Hops}
We further measure the model accuracy across different reasoning cases (hops) to better understand its reasoning capabilities (\autoref{tab:effect_ourCot_diff_reasoning}). First, aligned with the overall accuracy, our \apcot notably outperforms the conventional CoT (BLIP-2+CoT) in all reasoning cases on both VQA benchmarks (\eg, 1-hop in GQA: 43.88\% vs.~35.41\%, $\geq$2-hop in A-OKVQA: 48.09\% vs.~34.15\%).
We again attribute this to the benefit of incorporating the prediction into the CoT reasoning. Second, \apcot consistently improves over the baseline (BLIP-2) in all reasoning scenarios of both benchmarks, demonstrating the effectiveness of our answer reasoning path (\eg, 1-hop in A-OKVQA: 51.32\% vs.~46.70\%). 
More interestingly, on the challenging reasoning cases (\eg, 2 or more-hop reasoning), our \ktprompt and \apcot achieve notable gains over the baseline (\eg, GQA/A-OKVQA: 27.45\% vs.~7.66\% / 48.09\% vs.~46.54\%). 
In contrast, the traditional CoT achieves less gain or performs worse (GQA/A-OKVQA: 18.09\% / 34.15\%). This highlights the benefit of \ourapp, especially for solving complex reasoning questions. 

\mypar{Applicability of~\ourappbf} In addition to BLIP-2, we evaluate the effectiveness of our \ourapp on a more recent VLM, LLaVA-1.5~\cite{liu2023llava}. As shown in \autoref{tab:effect_ourCot_diff_reasoning}, \ourapp consistently improves over the traditional CoT in all reasoning cases, demonstrating its applicability to various VLMs with different architectural designs.

\subsection{Benefit of \ourappbf in fine-tuning stage}
\label{subsec:cot_finetune}
Besides assessing the zero-shot outcome, we evaluate \ourapp in the fine-tuning scenario. We first leverage the pre-trained BLIP-2 model with our \apcot to generate a rationale. 
We then use this generated rationale (together with the question and the image) to fine-tune the model on A-OKVQA (See the appendix for more details).
As shown in~\autoref{tab:fine_tuning}, our \apcot consistently improves the baseline (BLIP-2) over different reasoning cases, including multi-hop reasoning (\eg, $\geq$2-hop: 55.24\% vs.~54.92\%). This suggests the benefit of our \apcot even for fine-tuned models. Conversely, when the standard CoT is applied to BLIP-2 (BLIP-2+CoT), its performance notably degrades across all reasoning cases (\eg, $\geq$2-hop: 54.40\%), consistent with the findings from zero-shot experiments (\S\ref{subsec:cot_zershot}).
\begin{table}[t]
\centering
\small
\renewcommand\arraystretch{1.0}
\begin{tabular}{cccc}
\toprule
\multirow{2}{*}{Model} &
\multicolumn{3}{c}{A-OKVQA} \\
\cmidrule(r){2-4}
& 1-hop & $\geq$2-hop & All \\
\midrule
BLIP-2 & 57.63 & 54.92 & 56.88 \\
BLIP-2+CoT & 54.65 & 54.40 & 54.58 \\
BLIP-2+\apcot & \textbf{58.16} & \textbf{55.24} & \textbf{57.35} \\
\bottomrule
\end{tabular}
\caption{\small \textbf{Effectiveness of \apcotbf on A-OKVQA in the fine-tuning setting.} Aligned with the zero-shot results (\autoref{tab:effect_ourCot_diff_reasoning}), our \ourapp notably outperforms two baselines, BLIP-2 and BLIP2+CoT, on every reasoning case.}
\vspace{-10pt}
\label{tab:fine_tuning}
\end{table}

\begin{table}[t]
\centering
\small
\tabcolsep 8.5pt
\renewcommand\arraystretch{1.0}
\begin{tabular}{ccccc}
\toprule
\multicolumn{5}{c}{Hop Increase Percentage (\%)} \\
\cmidrule(r){1-5}
\multirow{2.5}{*}{GQA} & 0-hop & 1-hop & 2-hop & All \\
\cmidrule(r){2-5}
& 93.84 & 77.47 & 70.58 & 86.34 \\
\bottomrule
\end{tabular}
\caption{\small \textbf{Hop Increase Percentage by our augmentation.} We provide the LLM with the knowledge from large-scale text corpus (Wikipedia) and make existing questions more complex (\eg, 77.47\% of original 1-hop questions now have at least one more reasoning steps).}
\label{tab:hop_incr_perc_gqa}
\end{table}

\begin{table}[t]
\centering
\small
\renewcommand\arraystretch{1.0}
\begin{tabular}{ccccc}
\toprule
\multirow{2}{*}{Question} & \multicolumn{4}{c}{GQA} \\
\cmidrule(r){2-5}
Type & 0-hop & 1-hop & 2-hop & All \\
\midrule
Original Q & 49.62 & 42.41 & 7.66 & 44.63 \\
Augmented Q & 37.86 & 35.58 & 6.38 & 35.60 \\
\bottomrule
\end{tabular}
\caption{\small \textbf{Zero-shot accuracy on expanded GQA questions.} The performance on augmented questions (Augmented Q) notably declines against that on original questions (Original Q), suggesting increased reasoning in the expanded questions.}
\vspace{-10pt}
\label{tab:hop_incr_acc_gqa}
\end{table}

\subsection{Expanding questions with more reasoning}
\label{subsec:expand_que}
As depicted in \autoref{tab:gqa_aokvqa_hops}, the number of complex reasoning questions (\ie, 2-hop reasoning) is marginal in the GQA test-dev set, comprising only 3.73\%. We thus conduct an ablation study: increasing the number of hops for each original question using a large-scale knowledge-base (\eg, Wikipedia) and evaluating the model performance on these newly expanded questions. We first extract keywords from the original question and use them as queries to retrieve their relevant information from Wikipedia. We then input this extra information (with the original question) into the LLM to increase the reasoning complexity of the question.

\begin{figure}
    \small
    \centering
    \centerline{\includegraphics[width=1.0\linewidth]
    {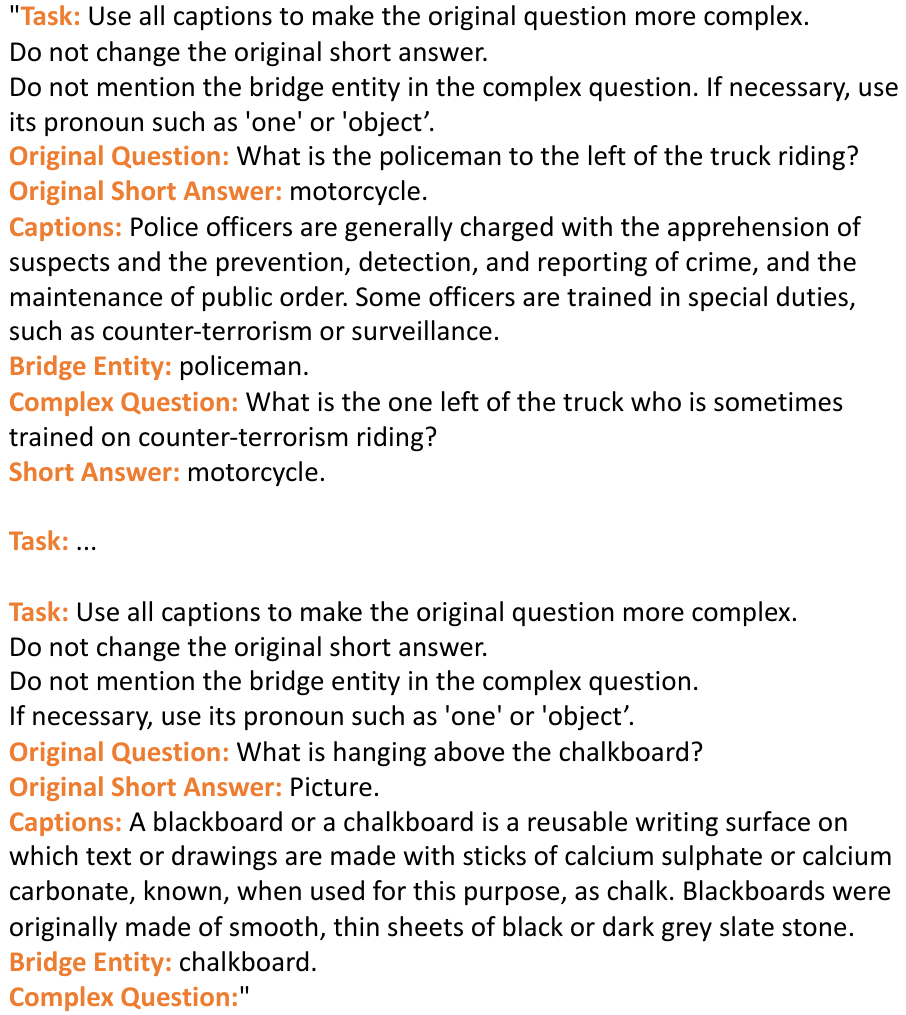}}
    \caption{\textbf{In-context language prompting to make the original question more complex.} ``Bridge Entity'' is a keyword extracted from the original question. ``Captions'' is the text snippet containing information about the bridge entity retrieved from Wikipedia. Using Wikipedia captions, we ask the LLM to increase the reasoning complexity in the original question while maintaining its original answer. We provide five in-context examples to the LLM.}
    \label{fig:aug_que_prompt}
\vspace{-10pt}
\end{figure}

\autoref{fig:aug_que_prompt} provides a detailed example of an in-context prompt for augmenting questions. Our primary objective is to increase the reasoning complexity of the original question while retaining its original answer. This enables us to conduct a more precise evaluation of how the model performance changes as the question becomes more intricate. We adopt a 5-shot in-context prompting where each example consists of seven components; ``Task'', ``Original Question'', ``Original Short Answer'', ``Captions'', ``Bridge Entity'', ``Complex Question'', and ``Short Answer''.

\begin{figure}
    \centering
    \centerline{\includegraphics[width=1.0\linewidth]{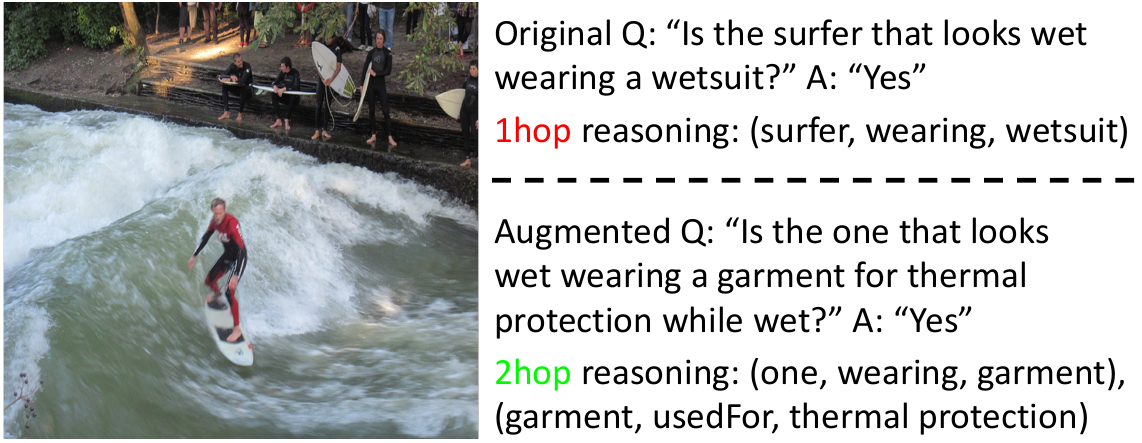}}
    \caption{\textbf{Qualitative results of augmenting question.} The original question now becomes more complex with one more reasoning step.}
    \label{fig:aug_que_qual}
\end{figure}

\autoref{fig:aug_que_qual} shows a qualitative example of augmenting the original question with more reasoning. The question originally requiring 1-hop reasoning (\ie, (``surfer'', ``wearing'', ``wetsuit)) now asks for 2-hop reasoning (\ie, (``one'', ``wearing'', ``garment''), (``garment'', ``usedFor'', ``thermal protection'')).

\autoref{tab:hop_incr_perc_gqa} shows that the number of hops in most original GQA questions has increased. For instance, 70.58\% of original 2-hop questions now require at least one more reasoning step than previously. In total, 86.34\% of original questions become more complex. 
We evaluate BLIP-2 on these new questions in the zero-shot setting (\autoref{tab:hop_incr_acc_gqa}). Compared to its performance on the original questions, we observe a notable drop in performance across all reasoning cases (\eg, 7.66\% vs.~6.38\% in the original 2-hop), indicating increased reasoning complexity in the newly expanded questions.

\begin{figure*}
    \centering
    \centerline{\includegraphics[width=1.0\linewidth]{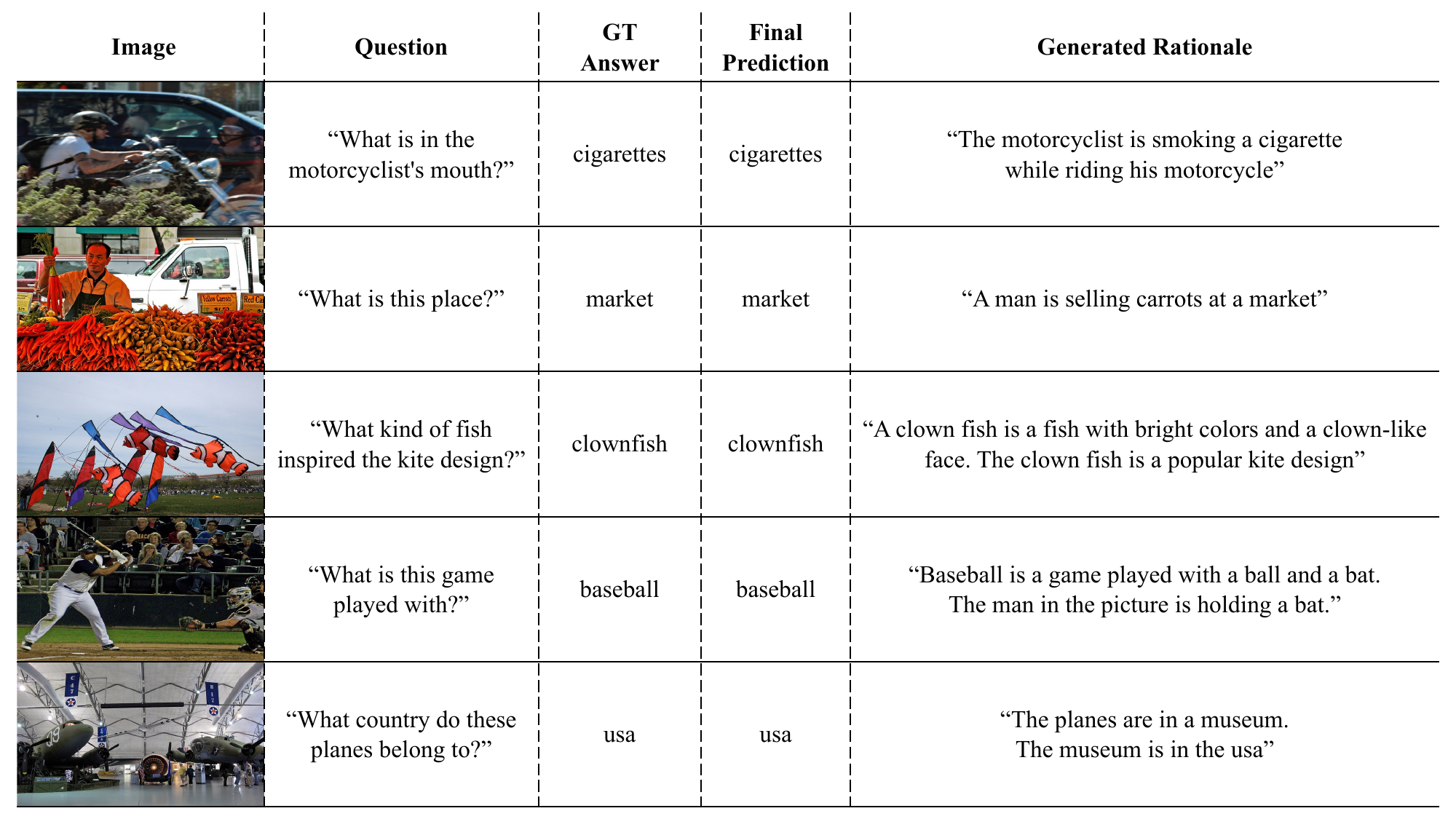}}
    \caption{\small \textbf{Qualtiative Results of \ourappbf.} Our rationales are highly relevant to the ground-truth answer. See details in \S\ref{subsec:qual_res}.}
    \label{fig:qual}
\end{figure*}

\subsection{Qualitative Results}
\label{subsec:qual_res}
\autoref{fig:qual} summarizes qualitative examples provided by our \apcot. The generated rationale is highly relevant to the correct answer, leading the model to make the correct prediction. Moreover, our rationales entail knowledge beyond images, such as commonsense (\eg, ``The clown fish is a popular kite design"), which is advantageous for A-OKVQA questions requiring external knowledge.

%% file: related.tex
\section{Related Work}
\label{sec:related}
\mypar{Multi-hop Reasoning}
Chain-of-thought (CoT) reasoning can be regarded as a form of multi-hop reasoning, as it involves constructing a sequence of reasoning to derive the answer. 
However, as indicated in~\autoref{tab:weak_trad_cot}, the traditional CoT method often results in erroneous information (\eg, visual hallucinations), leading to incorrect answers. Several recent works~\cite{dhuliawala2023chain,jiang2023active,chen2024measuring,zelikman2022star} have thus attempted to enhance the CoT reasoning capabilities. CoVe~\cite{dhuliawala2023chain} constructs a set of verification questions and uses them to verify the correctness of rationales. FLARE~\cite{jiang2023active} iteratively identifies flawed rationales and corrects them by utilizing relevant documents from the knowledge base. CURE~\cite{chen2024measuring} uses feedback from LLMs to tackle the hallucination during the generation process. 
Prophet~\cite{yu2023prophet} is related to our work as it first obtains answer candidates and then includes them in the LLM prompt for the final prediction. However, Prophet relies on separate models for answer candidate selection (VQA) and final prediction (GPT-3 \cite{Brown2020gpt3}), which is less flexible than our single \vl model approach. Additionally, Prophet makes direct final predictions without generating rationales, thus lacking multi-hop reasoning capabilities.
STaR~\cite{zelikman2022star} improves the model's reasoning capability by supervising rationale generation based on the inference results of the train sets. 
In contrast, our \ourapp proposes a self-correction mechanism,
which uses the initial prediction as the input context and autonomously corrects it if incorrect through the reasoning steps.


\mypar{Explaining Reasoning on VQA}
A few prior works~\cite{li2018vqa,wu2020improving,wu2019self,vaideeswaran2022towards} have explained the model's reasoning capabilities in the context of VQA tasks. VQA-E~\cite{li2018vqa} proposes a new VQA dataset derived from VQAv2 benchmark~\cite{goyal2017making} by synthesizing explanations for original VQAv2 samples. Some prior studies~\cite{wu2020improving,wu2019self} leverage human textual explanations to gain further insights into the model's reasoning abilities. More recently, \cite{vaideeswaran2022towards} aims to interpret the actions of VQA models by incorporating an end-to-end explanation generation module. Conversely, we utilize the LLM with novel language promptings grounded in the answer prediction and the knowledge triplet to automatically analyze various reasoning scenarios in VQA.

\mypar{Scene Graph and Knowledge Graph}
A scene graph (SG) from images and a knowledge graph (KG) related to questions~\cite{xie2022visual, singh2023coarse} are alternative ways to find the reasoning path to the answer. However, compared to rationales generated from LLM-based \vl models, SG and KG usually provide restricted visual semantic details in explaining the reasoning. This limitation arises as their graph generators~\cite{zheng2023prototype,schuster2015generating} often fail to capture diverse visual entities or semantic relations. 

%% file: conclusion.tex
\section{Conclusion}
We propose \ourapp to identify and improve multi-hop reasoning for VQA. \ourapp introduces two novel language promptings, an answer prediction-guided CoT prompt and a knowledge triplet-guided prompt, to generate a high-quality reasoning path to reach the answer. \ourapp utilizes this path to identify different reasoning scenarios in VQA benchmarks and consistently improves across all reasoning cases with a particular emphasis on complex reasoning questions.
\label{sec:conclusion}

%% file: limitations.tex
\section*{Limitations}
In this work, we propose \ourapp, a novel method to improve the reasoning capabilities of \vl models for VQA. We conduct a small-scale human study (involving 3 annotators) to assess the quality of our rationales. The evaluation may be subjective among annotators due to the nature of the language. For instance, each annotator may have different opinions about the number of reasoning steps required for the same question. We plan to expand the scale of the human study to mitigate this issue.

%% file: supp_content.tex
\appendix
\section*{Appendices}
In this appendix, we provide details omitted in the main text.
\begin{itemize} 
    [itemsep=5pt,topsep=1.5pt]
    \item \autoref{apdx:data_train}: More details about dataset and training (cf.~\S{\ref{sec:exp}}).
    \item \autoref{apdx:qual_incorr}: Qualitative results of incorrect prediction (cf.~\S{\ref{subsec:cot_zershot}}).
    \item \autoref{apdx:finetune}: Fine-tuning details (cf.~\S{\ref{subsec:cot_finetune}}).
    \item \autoref{apdx:add_res}: Additional Results (cf.~\S{\ref{sec:exp}}).
\end{itemize}

\begin{figure*}
    \centering
    \centerline{\includegraphics[width=1.0\linewidth]{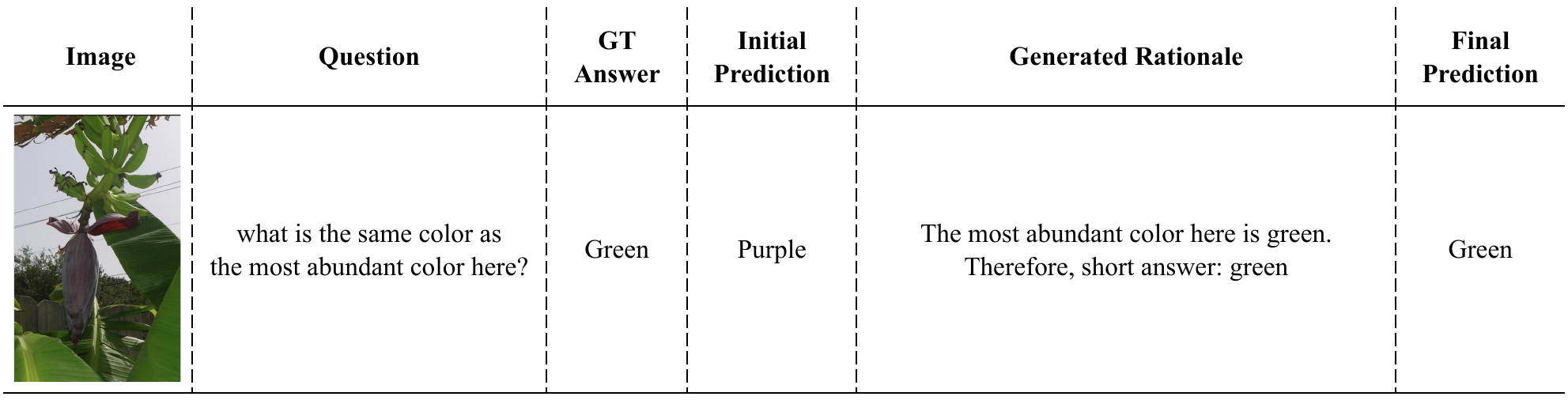}}
    \caption{\textbf{Qualtiative Results of our \ourappbf with incorrect prediction.}}
    \label{fig:qual_incorr_pred}
\end{figure*}

\section{More dataset and training details}
\label{apdx:data_train}
\mypar{Dataset}
We provide further details about the datasets used in our experiments.
GQA~\cite{gqa} is one of the popular VQA benchmarks comprising various visual compositional reasoning questions. GQA first obtains relations and attributes of visual objects from the scene graph~\cite{krishna2017visual} and utilizes them to generate VQA questions based on a pre-defined question engine. Followed by BLIP-2~\cite{blip2}, we use the official GQA train/test-dev splits for our experiments.
A-OKVQA~\cite{aokvqa} contains diverse VQA questions requiring real-world knowledge beyond the image, including commonsense and knowledge bases. Concretely, A-OKVQA has 25K questions, each offering both multiple-choice and direct-answer options. For our studies, we select the direct-answer option. We utilize the A-OKVQA train/validation splits, which consist of 17.1K/1.1K samples, respectively.

\mypar{Training}
We mainly use BLIP-2 as the \vl model for our experiments. Concretely, we opt for the largest BLIP-2 model, which features the ViT-g/14 vision encoder from EVA-CLIP~\cite{fang2023eva}, coupled with FlanT5-{XXL}~\cite{flant5}, an encoder-decoder-based LLM. BLIP-2 incorporates a transformer-based bridge module that connects the vision encoder to the LLM. We focus on training the LLM and the bridge component while keeping ViT frozen. Our training configuration entails a batch size of 16, a learning rate of 1e-5, a beam size of 5, a maximum sequence length of 512, and an image resolution of 490 for ten epochs. We utilize eight A100 48GB GPUs for both training and inference.

\section{Qualitative results of incorrect prediction}
\label{apdx:qual_incorr}
\autoref{fig:qual_incorr_pred} shows an example that our \apcot with the initial incorrect prediction leads the \vl model to make a correction, aligning with its quantitatively superior performance compared to the baseline (BLIP-2) (\eg, A-OKVQA All: 49.31\% vs.~46.05\% in \autoref{tab:effect_ourCot_diff_reasoning}).

\section{Fine-tuning details}
\label{apdx:finetune}
We provide details about fine-tuning our \apcot on A-OKVQA. As mentioned in~\S{\ref{subsec:cot_finetune}}, we initially utilize the same pre-trained BLIP-2 model used for zero-shot tasks, along with our \apcot, to generate an answer-related rationale. The main motivation for using the pre-trained model for rationale generation (instead of selecting a fine-tuned one on A-OKVQA) is that once the model undergoes fine-tuning, it loses its capability to generate rationales and shifts its primary focus to predicting answers directly. We thus deliberately select the pre-trained model for the effective rationale generation process. After obtaining the rationale, we prepend it (with the question and the image) to the answer-trigger prompt (\eg, ``Therefore, short answer:'') and fine-tune BLIP-2 on A-OKVQA using this prompt to predict the answer.

\section{Additional results}
\label{apdx:add_res}
\mypar{Accuracy of \apcotbf with the ground-truth answer as input context}
To accurately measure the number and the types of reasoning in A-OKVQA~\cite{aokvqa}, we utilize the ground-truth answer as the input context for \apcot, instead of initial answer prediction, which is the default setting. 
We observe a significant improvement in A-OKVQA performance when utilizing our \ourapp with the ground-truth answer compared to traditional CoT (69.34\% vs.~36.06\%), again supporting the high quality of our reasoning path for analyzing different reasoning scenarios (cf.~\S\ref{subsec:reasoning_stat}).